\documentclass[11pt]{article}

\usepackage{acl}

\usepackage{times}
\usepackage{latexsym}
\usepackage{amsfonts}
\usepackage{booktabs}
\usepackage{multirow}
\usepackage{array}
\usepackage{amssymb}
\usepackage{amsmath}
\usepackage[T1]{fontenc}

\usepackage[utf8]{inputenc}
\usepackage{enumitem}
\usepackage{microtype}

\usepackage{inconsolata}

\usepackage{graphicx}
\usepackage[table]{xcolor}
\definecolor{lightgray}{gray}{0.93}
\title{Addressing the Reasoning Gap: Mechanistic Circuit-Based Knowledge Editing in Large Language Models}

\author{
 \textbf{Tianyi Zhao\textsuperscript{1}},
 \textbf{Yinhan He\textsuperscript{1}},
 \textbf{Wendy Zheng\textsuperscript{1}},
 \textbf{Chen Chen\textsuperscript{1}}
\\
 \textsuperscript{1}University of Virginia
\\
\texttt{\{abs4dj,nee7ne,ncd9cf,zrh6du\}@virginia.edu}
}

\begin{document}
\maketitle
\begin{abstract}
Deploying Large Language Models (LLMs) in real-world dynamic environments raises the challenge of updating their pre-trained knowledge. While existing knowledge editing methods can reliably patch isolated facts, they frequently suffer from a \textit{reasoning gap}, where the model recalls the edited fact but fails to utilize it in multi-step reasoning chains.
Our analysis shows that multi-hop reasoning relies on sparse circuitry beyond direct factual recall, with functionally differentiated components contributing to intermediate processing and final-answer prediction.
To bridge this gap, we introduce MCircKE (\underline{M}echanistic \underline{Circ}uit-based \underline{K}nowledge \underline{E}diting), a mechanistically grounded framework that enables a map-and-adapt editing procedure.
MCircKE first identifies the causal circuits responsible for a specific reasoning task, capturing both the storage of the fact and the routing of its logical consequences. It then surgically update parameters exclusively within this mapped circuit.
Extensive experiments on the MQuAKE-series benchmarks demonstrate the effectiveness of the proposed method for multi-hop reasoning in knowledge editing.
\end{abstract}

\section{Introduction}
The deployment of large language models (LLMs) in dynamic, real-world environments is fundamentally constrained by the static nature of their pre-trained knowledge. Knowledge editing techniques~\cite{wang2024knowledge,zhang2024comprehensive} seek to address this limitation by modifying specific facts in a trained model without full retraining.
While recent methods can reliably update isolated single-hop facts, they often exhibit a critical failure mode termed the \textit{reasoning gap}~\cite{yao2025cake,zhanglocate}. In this phenomenon, a model may correctly recall an edited fact in isolation (e.g.,"Who is the PM?") but fail to propagate this updated information through multi-step reasoning (e.g., "Which political party is the PM the leader of?").
This discrepancy suggests that, although the storage of the fact has been successfully patched, the reasoning pathways required to utilize that fact remain misaligned.

Traditional knowledge editing methods~\cite{meng2022locating,wang2024wise,fangalphaedit} typically treat knowledge as isolated atomic units stored within specific multi-layer perceptron (MLP) layers. However, recent advances in mechanistic interpretability suggest that knowledge retrieval and utilization are instead governed by distributed reasoning circuits -- sparse computational subgraphs composed of attention heads and MLPs that route information responsible for performing a specific task across layers. From this perspective, this \textit{reasoning gap} is not merely a retrieval failure but a structural misalignment; the model’s logical wires are still connected to obsolete reasoning paths.
Recent work has begun to address this challenge. For instance, CaKE~\cite{yao2025cake} posits that the reasoning gap exists because standard edits do not compel the model to practice using the new knowledge. It mitigates this by generating circuit-aware synthetic multi-hop data derived from the edited fact, and fine-tuning the model on this data. By training on this curated data, CaKE aims to implicitly activate the dynamic reasoning circuits needed for generalization. While effective, this strategy relies on the indirect pressure of data augmentation to realign the circuits, and thus still treats the internal mechanisms as a black box that is expected to self-correct given appropriate inputs.


\begin{figure*}[t]
\centering
\includegraphics[width=\linewidth]{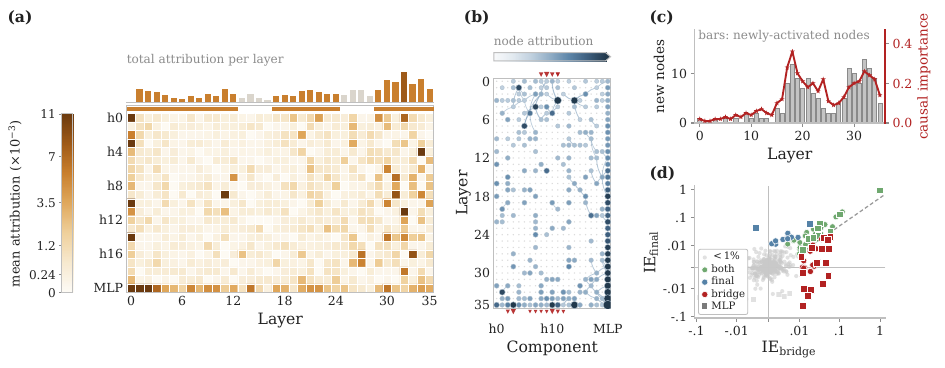}
\caption{{Mechanistic analysis of multi-hop factual reasoning in GPT-2 Large.}
\textbf{(a)} Aggregate heatmap of attribution scores across 100 instances reveals distinct layer-wise computational regimes.
\textbf{(b)} A sparse causal subgraph for a representative two-hop instance reveals a layer-localized routing pathway rather than a dense activation pattern.
\textbf{(c)} Gray histogram shows the number of nodes active in the multi-hop circuit but inactive in the single-hop circuit; the red curve shows their aggregated causal importance. 
The clear mid-to-late-layer surge reveals additional important components that remain dormant during single-hop retrieval.
\textbf{(d)} Causal ablation distinguishes routing/bridge heads from destination heads by their effects on the intermediate entity and final answer, respectively.
}
\label{fig:panel}
\end{figure*}

To tackle the aforementioned issue, we propose MCircKE, a mechanistically grounded framework that addresses the limitations of existing knowledge editing methods and fundamentally diverges from the data-stimulation strategy of CaKE.
Rather than relying on pre-constructed data to implicitly activate reasoning paths, our approach aims to explicitly and precisely identify the dynamic reasoning circuits responsible for the target knowledge.
Specifically, we first identify a rigorous, high-fidelity subgraph of the edges and nodes that causally contribute to the model’s reasoning process. By integrating gradients along the activation path, we can pinpoint the exact wires in the neural network that carry the knowledge to be edited.
Once this reasoning circuits are identified, we then surgically adapt the model along the mapped paths.
This transforms knowledge editing from a heuristic "locate-and-replace" operation into a structural "map-and-adapt" procedure. By explicitly targeting components that have been verified as part of the reasoning chains, our method ensures that the updated knowledge is not only stored but also mechanistically integrated into the model’s logic.
In summary, our main contributions are:
\begin{itemize}[itemsep=-3pt,topsep=-2pt]
    \item \textbf{Mechanistic Insight}: We conduct a mechanistic analysis of multi-hop factual reasoning in LLMs, providing coarse-to-fine-grained insights into why existing knowledge editing methods fail at multi-hop factual recall and how this failure can be addressed.
    \item \textbf{MCircKE Framework}: We propose a mechanistically grounded editing framework that explicitly maps dynamic reasoning circuits and performs surgical, path-constrained updates.
    \item \textbf{Empirical Validation}: Extensive experiments demonstrate that MCircKE substantially improves multi-hop reasoning performance for knowledge editing in LLMs.
\end{itemize}


\section{Preliminaries}
\label{sec:preliminaries}
\subsection{LLM as a Computational Graph}
We represent an LLM as a directed acyclic computational graph $\mathcal{G}=(\mathcal{V},\mathcal{E})$, where nodes $\mathcal{V}$ correspond to Transformer submodules, including attention projections and MLPs, and edges $\mathcal{E}$ represent information flow between modules through the residual stream. Specifically, an edge $e_{u\rightarrow v}$ exists when the output of module $u$ contributes to the input of module $v$. This representation enables reasoning behavior to be localized at a finer granularity than entire Transformer layers. A detailed definition is provided in App.~\ref{app:eapig}.

\subsection{Edge Attribution with EAP-IG}
To identify the subgraph responsible for a target behavior, we use Edge Attribution Patching with Integrated Gradients (EAP-IG)~\cite{hanna2024have}. EAP-IG estimates the importance of each edge by integrating gradients along a path from a corrupted input $I_{\mathrm{corrupt}}$ to the clean input $I_{\mathrm{clean}}$, alleviating the gradient-saturation issue of vanilla EAP~\cite{nanda2023attribution}. For an edge $e$, its attribution score is approximated as \begin{equation}\label{eq:approx} \phi(e) \approx (x_e^{\mathrm{clean}}-x_e^{\mathrm{corrupt}}) \frac{1}{m}\sum_{k=1}^{m} \frac{\partial \mathcal{L}(x(k/m))}{\partial x_e}, \end{equation} where $m$ is the number of integration steps.

\noindent\textbf{Causal circuit.} 
We define the causal circuit $\mathcal{C}(K)\subset\mathcal{G}$ as the sparse subgraph induced by the top-$K$ edges ranked by attribution magnitude: \begin{equation}\label{eq:circuit-def} \mathcal{C}(K) = \mathrm{top\text{-}}K \left\{ e\in\mathcal{E}\mid |\phi(e)| \right\}. \end{equation} We use this circuit as the mechanistic unit for subsequent analysis and editing. Further details on EAP-IG and its computation are provided in App.~\ref{app:eapig2}.

\section{Mechanistic Analysis}\label{sec:analysis}
We start by conducting a mechanistic diagnostic study to uncover the causal circuitry supporting multi-hop reasoning and to understand, at a structural level, why successful single-hop edits may fail to generalize to multi-hop inference.
Figure~\ref{fig:panel} presents a coarse-to-fine analysis from layer-wise attribution to instance-level circuitry and causal functional specialization.

\subsection{Inter-Layer Distribution of Causal Importance}

\noindent\textbf{Setup.} We randomly sample 100 two-hop instances from MQuAKE-3K-v2~\cite{zhong2023mquake}, compute an EAP-IG circuit for each, and aggregate component-level attribution scores. 
Figure~\ref{fig:panel}(a) thus captures the general layer-wise structure of multi-hop reasoning in GPT-2 Large.

\noindent\textbf{Layer-wise Computational Structure.}
Three broad regimes emerge.
Early layers (0--10) are dominated by MLP attribution, consistent with prior findings established in ROME~\cite{meng2022locating} and MEMIT~\cite{mengmass}: these layers are responsible for the initial retrieval of the atomic fact. Standard methods usually successfully target these layers.
%
Crucially, there is a distinct cluster of activity in MLPs around layers 17--22, possibly suggesting processing during compositional inference.
Late layers (29--35) show strong activations of both attention heads and MLPs, indicating further information routing and processing.


\subsection{Intra-Layer Sparsity and Differential Circuit Composition}\label{sec:intra-layer}
The above aggregated view captures the general layer-wise trend. To complement it, we now zoom into a single representative two-hop instance drawn from MQuAKE-3K-v2, \texttt{Crysis\_2$\rightarrow$Crytek$\rightarrow$English}, and compare its circuit with the corresponding single-hop circuit \texttt{Crytek$\rightarrow$English}.

\noindent
\textbf{Visual Analysis of Sparse Circuitry.}
Figure~\ref{fig:panel}(b) shows the circuit extracted for \texttt{Crysis\_2$\rightarrow$Crytek$\rightarrow$English}, where only a small subset of heads is active, forming a sparse, layer-localized pathway with instance-specific structure not fully captured by the aggregate pattern. Even within highly attributed layers, participation is concentrated in a few heads rather than uniformly distributed. This intra-layer sparsity suggests that layer-level editing may be overly coarse and unnecessarily affect unrelated components and functions.

\noindent
\textbf{Differential Activation Dynamics.}
We further compare the multi-hop chain \texttt{Crysis\_2$\rightarrow$Crytek$\rightarrow$English} with its corresponding single-hop circuit \texttt{Crytek$\rightarrow$English}. We define {newly activated nodes} as components active in the multi-hop circuit but inactive in the single-hop circuit. As shown in Figure~\ref{fig:panel}(c), these components concentrate in the mid-to-late layers and carry substantial aggregate attribution, revealing a clear shift in circuit composition beyond direct factual recall. Thus, edits localized from single-hop traces may update the target association while overlooking computation additionally engaged during multi-hop reasoning.

\begin{figure}[h]
\centering
\includegraphics[width=0.88\linewidth]{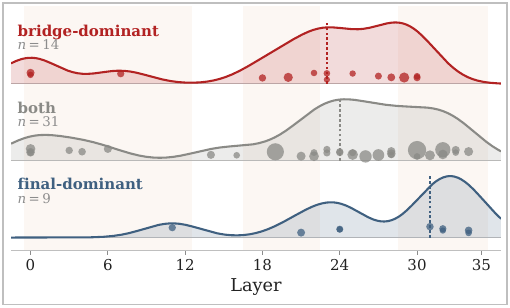}
\caption{\textbf{Layer distribution of attention-head roles.}
Points denote heads with non-negligible indirect effects; curves show within-role layer distributions and dashed lines their medians.}
\label{fig:headrole}
\end{figure}

\subsection{Functional Characterization of Attention-Head Roles}\label{ana3}
The preceding analysis shows that multi-hop reasoning engages additional mid-to-late circuitry beyond direct factual recall, but does not reveal the roles of these components. We therefore perform head-level causal intervention over 100 cases and measure, for each component, its normalized indirect effect on the intermediate bridge entity, $\mathrm{IE}_{\mathrm{bridge}}$, and on the final answer, $\mathrm{IE}_{\mathrm{final}}$~\cite{pearl2022direct,vig2020investigating}.
Figure~\ref{fig:panel}(d) reveals distinct causal profiles. Among components with non-negligible effects, some preferentially restore the bridge entity ({bridge-dominant}), others preferentially restore the final answer ({final-dominant}), while many contribute to both.
The corresponding layer distributions in Figure~\ref{fig:headrole} further show that bridge-dominant heads tend to occur earlier than final-dominant heads. These results support a differentiated computation in which intermediate information processing and final-answer computation rely on partially distinct components.
Additional details are presented in Appendix~\ref{app:mech_func}.

\paragraph{Summary.}
The above analysis provides the mechanistic justification for circuit-based editing. 
Effective multi-hop editing requires not only updating the factual association itself, but also adapting the routing pathways that propagates and integrates the edited information into downstream inference.

\section{MCircKE: Mechanistic Circuit-based Knowledge Editing for LLMs}
Building on the insights from our mechanistic analysis, we introduce MCircKE (\underline{M}echanistic \underline{Circ}uit-based \underline{K}nowledge \underline{E}diting), a framework designed to bridge the reasoning gap by editing the specific causal circuits responsible for knowledge utilization.
Unlike heuristic methods that target fixed layers or data-driven methods that rely solely on implicit stimulation, our approach explicitly maps the causal reasoning pathway for a specific fact and surgically adapts the model along that pathway.


\subsection{Circuit Discovery}
Our first objective is to identify the reasoning circuits $\mathcal{C} \subset \mathcal{G}$ casually responsible for the target reasoning task.

\begin{figure}[h]
\centering
\includegraphics[width=\linewidth]{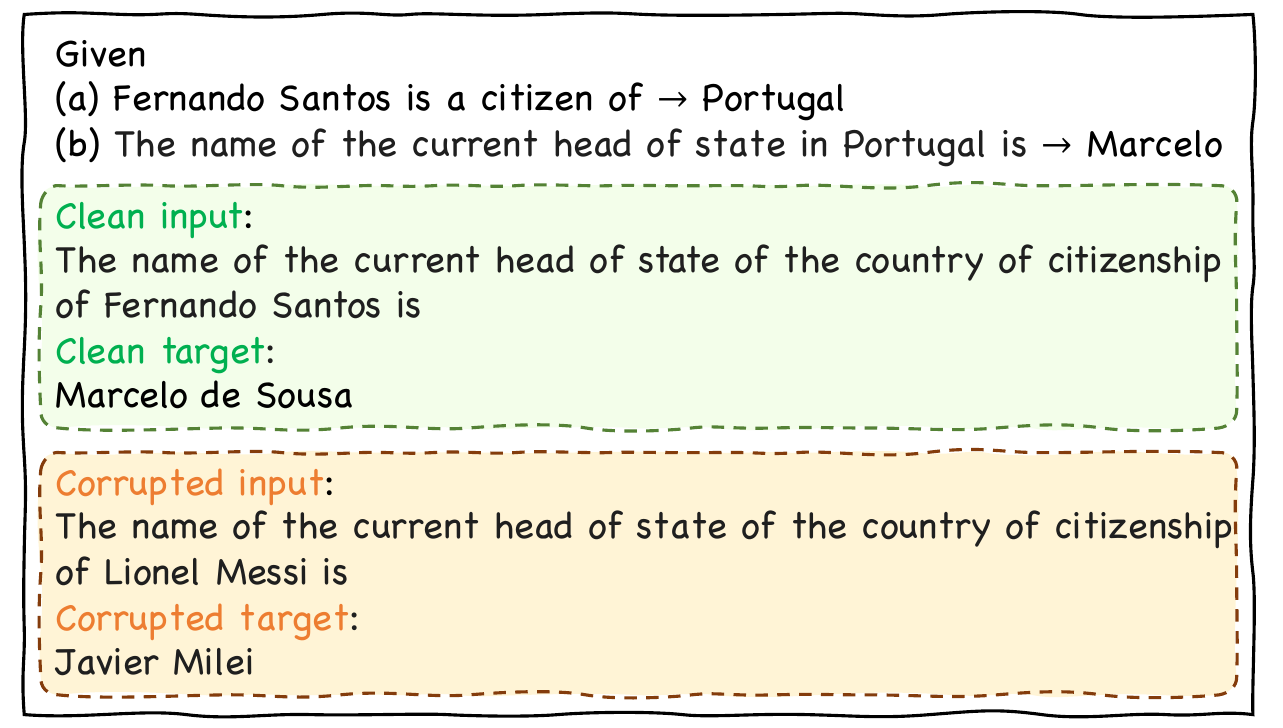}
\caption{Example clean and corrupted inputs.}
\label{fig:example}
\end{figure}

\paragraph{Clean and Corrupted Input Pairs}
\textit{Clean prompts construction}: We first combine the provided single-hop statements from the original knowledge and chain them into a unified multi-hop prompt.
\textit{Corrupted prompts construction}: We then utilize ChatGPT to generate a corresponding corrupted prompt that maintains semantic manifold consistency to the clean prompt. This ensures the prompt structure and reasoning complexity remain identical, but the subject entity is altered to an alternative.
An example is shown in Figure~\ref{fig:example}.

\paragraph{Attribution Integration}
We compute the EAP-IG score $\phi (e)$ for all edges in the model by integrating the gradients along the linear interpolation path between the corrupted and clean embeddings, using the batched Riemann approximation (Equation~\ref{eq:approx}) with $m=5$.
We use the magnitude $|\phi (e)|$ because we are interested in causal relevance, regardless of whether the edge positively or negatively impacts the output.

\begin{table*}[t]
\centering
\begin{tabular}{c l c c c c c}
\toprule
\multirow{2}{*}{Model}
& \multirow{2}{*}{Method}
& \multirow{2}{*}{Overall M-Acc.}
& \multicolumn{3}{c}{\#hops}
& \multirow{2}{*}{S-Acc.} \\
\cmidrule(lr){4-6}
& & & 2-hop & 3-hop & 4-hop & \\
\midrule
\multirow{8}{*}{\rotatebox{90}{GPT-2 Large}}
& LoRA        & 24.0858 & 39.3886 & 13.4390 & 16.8724 & 48.947 \\
& ROME        & 14.9939 & 17.2546 & 13.1162 & 14.4033 & 52.956 \\
& MEMIT       &  7.7804 & 10.9053 &  5.7805 &  6.0357 & 35.009 \\
& WISE        &  8.1138 & 11.6990 &  5.9272 &  5.9442 & 21.421 \\
& AlphaEdit   &  8.5028 & 13.2275 &  5.6925 &  5.5327 & 44.093 \\
& CaKE        & 34.9561 & 37.0370 & 29.3721 & 40.4207 & \underline{71.666} \\
\cmidrule(lr){2-7}
& \textbf{MCircKE}
              & \textbf{50.4168}
                          & \textbf{67.9306}
                                      & \underline{38.6737} & \underline{41.4723} & 64.915 \\
& \textbf{MCircKE}$^{\bigstar}$
              & \underline{47.2824} & \underline{54.1446} & \textbf{39.1432} & \textbf{49.2913} & \textbf{73.213} \\

\midrule

\multirow{8}{*}{\rotatebox{90}{GPT-2 XL}}
& LoRA        & 25.6085 & 41.9753 & 15.0528 & 16.5981 & 52.933 \\
& ROME        & 12.5042 & 17.7543 &  9.8298 &  8.5048 & 61.511 \\
& MEMIT       &  8.4806 & 12.5220 &  5.7218 &  6.4929 & 46.759 \\
& WISE        &  8.3917 & 11.7578 &  5.7512 &  7.2702 & 22.876 \\
& AlphaEdit   &  9.4587 & 14.6972 &  6.0446 &  6.6301 & 48.749 \\
& CaKE        & 40.5246 & 46.6196 & 33.3040 & 42.2954 & \underline{76.659} \\
\cmidrule(lr){2-7}
& \textbf{MCircKE}     & \underline{49.0140} & \textbf{64.7860} & \underline{39.0020} & \underline{45.4960} & 73.894 \\
& \textbf{MCircKE}$^{\bigstar}$
              &   \textbf{49.3536}    &   \underline{58.1423}    &  \textbf{39.7887}     & \textbf{53.2468}      & \textbf{77.446} \\

\bottomrule
\end{tabular}
\caption{Performance comparison under different hop settings on GPT-2 Large and GPT-2 XL.}
\label{tab:result}
\end{table*}

\paragraph{Circuit Pruning}
The raw attribution map assigns a non-zero score to almost every edge. To obtain the sparse reasoning circuits, we employ top-$K$ pruning. Specifically, we rank all edges based on the magnitude of their attribution scores $|\phi (e)|$, and retain only the top $K$ edges with the highest attribution scores, effectively filtering out noise and irrelevant pathways.

\subsection{Circuit-Guided Low-Rank Adaptation}
Once the circuits $\mathcal{C}$ are identified, we apply Low-Rank Adaptation exclusively to the modules within the circuits.
For every weight matrix $W \in \mathcal{C}$ identified in the reasoning circuit, we freeze $W$ and introduce low-rank matrices $A$ and $B$:
\begin{equation}
    W_{new} = W + \Delta W = W + \frac{\alpha}{r} B A
\end{equation}
where $B \in \mathbb{R}^{d \times r}$, $A \in \mathbb{R}^{r \times k}$. Here $\alpha$ is a scaling factor and $r$ is the rank, with $r \ll d$.
Critically, for any module $W' \notin \mathcal{C}$ (parameters outside the reasoning circuits), we enforce $\Delta W' = 0$. This constraint ensures that the edit is mechanistically localized and physically rewires the specific information flow responsible for the logic.

\section{Experiments}
In this section, we aim to answer the following research questions:
\begin{itemize}[itemsep=-3pt,topsep=-2pt]
    \item \textbf{RQ1}: Can the proposed method effectively bridge the reasoning gap in multi-hop factual recall compared to baselines?
    \item \textbf{RQ2}: Does the proposed method ensure the preservation of unrelated knowledge?
    \item \textbf{RQ3}: Is precise topological identification of reasoning circuits necessary, and do the identified circuits demonstrate validity relative to stochastic and heuristic baselines?
\end{itemize}
\subsection{Evaluation Setup}
\paragraph{Dataset.}
We conduct the evaluation on the challenging MQuAKE-3K-V2 benchmark (Multi-hop Question Answering for Knowledge Editing)~\cite{zhong2023mquake}, a widely used dataset for evaluating multi-hop factual recall with 3,000 counterfactual editing instances.

\paragraph{Baselines and Models.}
We compare our method against several representative knowledge editing baselines, including: 
\textbf{LoRA}~\cite{hu2022lora}, which directly fine-tunes the full model;
\textbf{ROME}~\cite{meng2022locating}, a classic locate-then-edit approach;
\textbf{MEMIT}~\cite{mengmass}, an extension of ROME that performs edits across a range of early layers;
\textbf{WISE}~\cite{wang2024wise}, which augments the model with a side memory and edits later layers;
\textbf{AlphaEdit}~\cite{fangalphaedit}, which updates parameters via projection into a null space;
\textbf{CaKE}~\cite{yao2025cake}, which constructs additional multi-hop edit prompts to improve multi-hop factual recall.
We conduct experiments on three GPT-2 variants~\cite{radford2019language}: GPT-2 Large (774M), GPT-2 XL (1.5B), and GPT-J (6B); and additionally evaluate at 7B scale on Qwen2.5-7B~\cite{qwen2025qwen25technicalreport}\footnote{Addtional results in App.~\ref{app:exp_more}}.

\paragraph{Metrics.}
We assess the model’s ability to leverage edited knowledge along three key dimensions: \textit{Multi-hop Accuracy}, measuring performance on questions requiring two-, three-, and four-hop reasoning over the edited fact; \textit{Single-hop Accuracy}, capturing direct recall of the edited knowledge; and \textit{Locality}, quantifying the extent to which an edit preserves the model’s behavior on queries unrelated to the targeted knowledge.

\paragraph{Implementation.}
We evaluate our proposed method in two configurations to isolate the impact of data versus structure:
\underline{MCircKE}, which edits the model using only the single-hop edit sample for the target fact provided in MQuAKE;
\underline{MCircKE$^{\bigstar}$}, which additionally incorporates synthetic multi-hop training data from CaKE~\cite{yao2025cake} for model editing.
For the hyperparameters, we set $K=4{,}000$ for GPT-2 Large, $K=10{,}000$ for GPT-2 XL, and $K=5{,}000$ for GPT-J. $\alpha$ is set to $8$ and rank $r$ is set to $32$ in the circuit-guided low-rank adaptation.
We run our experiments on 6 NVIDIA A6000 GPUs.

\subsection{Experimental Results}

\paragraph{Compared with baselines, MCircKE consistently bridges the reasoning gap across all model scales.}
Table~\ref{tab:result} presents the multi-hop and single-hop accuracies for all methods on GPT-2 Large and GPT-2 XL. We also present the comparative performance on GPT-J in Table~\ref{tab:gptj}.
Overall, MCircKE achieves the highest multi-hop accuracy among all methods, delivering a substantial improvement over the best previous approach.
For instance, on GPT-2 Large, our method attains an overall multi-hop accuracy of 50.4\%, outperforming CaKE (which achieves 34.9\%) by over 15 percentage points. On GPT-2 XL, MCircKE outperforms the strongest baseline by approximately 9\% in overall multi-hop performance. On GPT-J, it reaches 59.25\% overall multi-hop accuracy, again surpassing the best baseline.
Notably, the gains are consistent across different reasoning depths.
For instance, on GPT-2 XL, it yields consistent gains on 2-hop questions (49.35\% vs. 40.52\% for CaKE), 3-hop (39.78\% vs. 39\%) and 4-hop queries (53.24\% vs. 45.49\%).
These results provide strong evidence in support of RQ1, demonstrating that our method effectively enhances multi-hop factual recall and enables the model to answer complex, chained queries grounded in the edited knowledge.

\begin{table}[t]
\centering
\resizebox{\linewidth}{!}{%
\begin{tabular}{l c c}
\toprule
Method 
& M-hop (2-hop / 3-hop / 4-hop) 
& S-hop \\
\midrule
LoRA    
& 9.769 (12.61 / 6.72 / 10.11) 
& 47.579 \\
ROME    
& 23.23 (32.89 / 17.55 / 17.06) 
& 52.497 \\
CaKE    
& \underline{56.318} (\underline{61.84} / \underline{46.77} / \underline{62.59}) 
& \textbf{85.277} \\
\midrule
\textbf{MCircKE} 
& \textbf{59.253} (\textbf{64.61} / \textbf{49.41} / \textbf{66.26}) 
& \underline{85.219} \\
\bottomrule
\end{tabular}
}\caption{Performance comparison under different hop settings on GPT-J.}
\label{tab:gptj}
\end{table}

\paragraph{The collapse of unstructured fine-tuning.}
Standard LoRA struggles significantly compared to MCircKE. For instance, on GPT-2 XL, LoRA achieves only $\sim$25\% multi-hop accuracy compared to MCircKE's $\sim$49\%. 
This gap suggests that, with limited topological guidance, global parameter updates must search an extremely large parameter space to recover the correct reasoning pathways, making it fail to converge to the desired circuit-level change. In contrast, MCircKE’s strong performance indicates that mechanistic circuit guidance is crucial for reliable knowledge editing with multi-hop generalization.

\paragraph{The reasoning vs. recall trade-off.}
Comparing the standard MCircKE with CaKE (trained on addtional multi-hop data) reveals a critical trade-off.
MCircKE consistently outperforms CaKE on multi-hop reasoning (e.g., 50.42\% vs. 34.96\% accuracy on GPT-2 Large), indicating that repairing circuit structure yields stronger generalization than implicit behavioral imitation.
However, CaKE retains a slight edge in single-hop recall. This suggests that CaKE's approach of fine-tuning on augmented data leads to stronger surface-level memorization of the facts, whereas MCircKE's surgical approach prioritizes the logical consistency of the circuit, occasionally at the cost of absolute recall strength for the atomic fact itself.

\begin{table}[h]
\centering
\begin{tabular}{l c c}
\toprule
$\left|\Delta\right|$ (\%) & MMLU & CSQA \\
\midrule
LoRA      & 0.34 & 1.24 \\
ROME      & 0.06 & 0.41 \\
AlphaEdit & 0.21 & 0.25 \\
CaKE      & 0.09 & 0.49 \\
MCircKE   & 0.10 & 0.24 \\
\bottomrule
\end{tabular}
\caption{Locality performance.}
\label{tab:locality}
\end{table}

\paragraph{The failure of heuristic "Locate-and-Edit".}
Traditional methods like ROME and MEMIT perform poorly on multi-hop reasoning (M-Acc < 15\% on GPT-2 Large and GPT-2 XL).
While they achieve decent Single-hop Accuracy, they fail to integrate this knowledge into the reasoning chain. This confirms our mechanistic hypothesis derived in Section~\ref{sec:analysis}: fixing the "storage" without fixing the "routing"  leaves the model unable to leverage the new fact.
ROME effectively creates "orphaned facts" that can be recalled but not reasoned with, and the resulting reasoning gap (delta between S-Acc and M-Acc) is stark.
On GPT-2 XL, ROME exhibits a $\sim$49\% drop from single-hop to multi-hop accuracy (61.5\% S-Acc vs. 12.5\% M-Acc). In contrast, MCircKE shrinks this gap to $\sim$25\% while achieving higher S-Acc and M-Acc, demonstrating significantly better knowledge integration.

\paragraph{Locality and stability analysis.}
To assess locality, Table~\ref{tab:locality} presents the performance change on two unrelated benchmarks, MMLU~\cite{hendrycks2021measuring} and CommonsenseQA~\cite{talmor-etal-2019-commonsenseqa}, which evaluate the model’s general abilities after applying the knowledge editing methods.
Overall, the minimal performance shifts indicate that MCircKE’s edits are highly localized, integrating new knowledge without disrupting the model’s existing unrelated knowledge.

\subsection{Sensitivity Analysis}
\begin{figure}[h]
\centering
\includegraphics[width=0.8\linewidth]{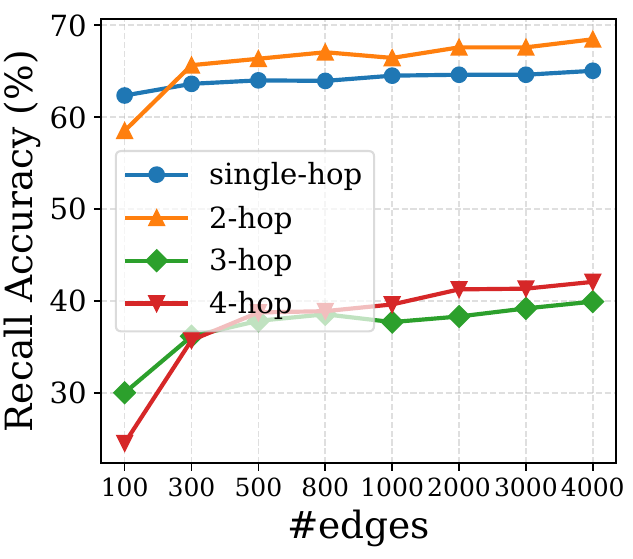}
\caption{Effect of the number of edges retained in the discovered circuit graphs.}
\label{fig:e}
\end{figure}
\noindent
A core hyperparameter in MCircKE is the number of edges retained in the reasoning circuit. A very sparse circuit is desirable for locality (modifying fewer parameters), but an overly sparse circuit may sever critical reasoning pathways.
Figure~\ref{fig:e} illustrates the impact of varying the edge count from 100 to 4000 on GPT-2 Large.
We can observe that:
\begin{itemize}[itemsep=-3pt,topsep=-2pt]
    \item At extreme sparsity levels (100-300 edges), single-hop accuracy remains relatively robust ($\sim$62\%), but multi-hop performance suffers catastrophic failure (4-hop accuracy drops to $\sim$25\%). This disparity suggests that simple factual recall relies on a highly localized set of "storage" edges (likely within specific MLP layers), whereas multi-hop reasoning requires a more extensive "routing" infrastructure.
    \item As the edge count increases, we observe that performance effectively plateaus beyond 2,000 edges. This saturation point suggests that the logic required to process a specific fact is not diffuse across the entire model but is concentrated in a fraction of the network.
\end{itemize}

\subsection{Ablation Study: Circuit Validity}
To further verify the effectiveness of editing the model along the identified circuits, we conducted an ablation study on 1,000 cases from the MQuAKE dataset using GPT-2 Large. We compared MCircKE against three alternative strategies:
\textbf{Full Graph} - Updating all edges in the model (equivalent to standard full model fine-tuning).
\textbf{Only MLPs} - Updating all MLP modules while freezing all other modules.
\textbf{Random Edges} - Updating a random subset of edges equal in size to the MCircKE circuits.
The results are shown in Table~\ref{tab:abla}.
We can observe that:

\begin{table}[t]
\centering
\begin{tabular}{l c c}
\toprule
Method & Single-hop & Multi-hop \\
\midrule
Full Graph    & 61.40 & 39.30 \\
Only MLPs     & 58.85 & 26.00 \\
Random Edges  & 26.35 & 15.13 \\
\rowcolor{lightgray}
\textbf{MCircKE} & \textbf{73.17} & \textbf{67.73} \\
\bottomrule
\end{tabular}
\caption{Ablation Results.}\label{tab:abla}
\end{table}

\begin{itemize}[itemsep=-3pt,topsep=-2pt]
    \item MCircKE drastically outperforms Random Edges, confirming that the circuits utlized in the proposed method are mechanistically relevant and not merely a random sub-network.
    \item The Only MLPs baseline illustrates the precise failure mode of traditional editing. While it achieves decent Single-hop accuracy, its Multi-hop performance collapses to 26.00\%. This empirically validates the hypothesis that modifying storage alone is insufficient; without the concurrent update of routing circuits, the new knowledge cannot be propagated.
    \item MCircKE outperforms the Full Graph update. This suggests that restricting updates to the relevant circuits acts as a form of structural regularization. By freezing irrelevant pathways, we prevent the noise of global updates from interfering with the delicate logic of the reasoning chains, allowing the focus on rewiring the specific factual bridge.
\end{itemize}

\begin{figure}[h]
\centering
\includegraphics[width=0.98\linewidth]{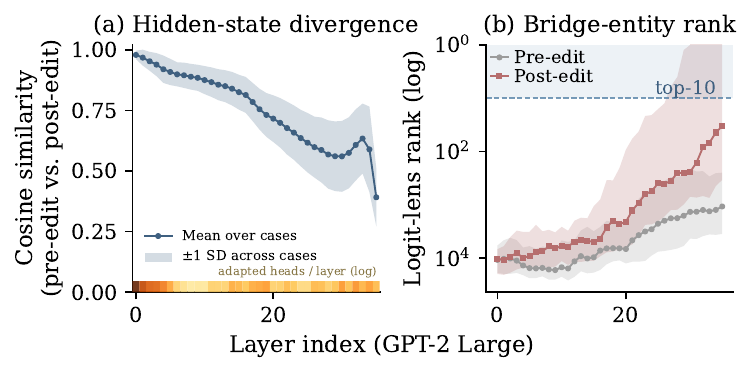}
\caption{\textbf{Representation-level effects of editing.}
(a) Editing induces progressively larger hidden-state shifts across depth.
(b) First-hop edits substantially elevate the counterfactual bridge entity in logit-lens rank.
}
\label{fig:rep}
\end{figure}

\subsection{Pre-/Post-Edit Internal Representation Analysis}
We ask whether MCircKE induces mechanistically meaningful changes in internal representations, examining two complementary signals: how hidden states evolve across depth after editing, and whether the counterfactual bridge entity rises in rank across layers under the logit lens.

\noindent\textbf{Hidden-State Divergence.}
Figure~\ref{fig:rep}(a) shows that pre/post hidden-state similarity decreases across depth, with stronger divergence in the middle-to-late layers.
This trend is qualitatively consistent with the mid-to-late computational regimes identified in our mechanistic analysis.

\noindent\textbf{Bridge-Entity Tracking.}
Figure~\ref{fig:rep}(b) traces the counterfactual bridge entity through a layer-wise logit lens for first-hop edits. Its post-edit rank improves markedly across depth, reaching a median of $34$ at the final layer compared with $1072$ before editing. This trajectory shows that MCircKE promotes the edited bridge entity within the model's evolving representation, providing additional evidence that the new fact is integrated into the intermediate reasoning process rather than merely recovered at the output.

\begin{figure}[h]
\centering
\includegraphics[width=0.85\linewidth]{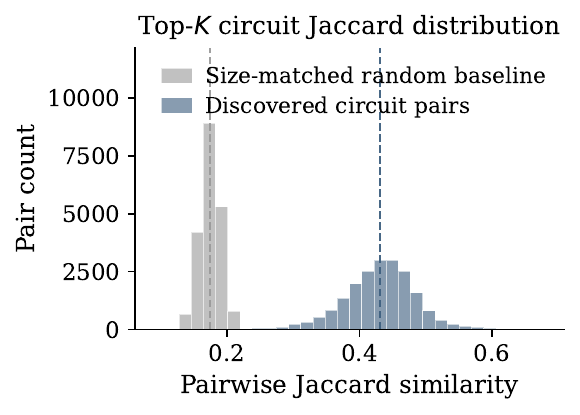}
\caption{{Cross-instance circuit consistency.}}
\label{fig:consist}
\end{figure}
\subsection{Discussion: Toward Reusable Circuits and Amortized Editing}

A practical concern for per-fact circuit editing is cost: if a new circuit must be discovered for every fact, the overhead may be difficult to amortize. This motivates a natural question: to what extent are discovered circuits shared across instances?

To examine this, we analyze cross-instance circuit consistency over 200 GPT-2 Large MQuAKE-CF-3K circuits with $K=3000$. Treating each circuit as a component set $C_i \subseteq U$, we measure pairwise overlap using the Jaccard coefficient,
\begin{equation}
J(C_i,C_j)=\frac{|C_i\cap C_j|}{|C_i\cup C_j|},
\end{equation}
over all $19{,}900$ circuit pairs. We compare against a size-matched random baseline that samples two random component subsets with the same cardinalities as the observed pair. 
%
%
As shown in Figure~\ref{fig:consist}, discovered circuits exhibit substantially greater overlap than random subsets, indicating a strong cross-instance shared structure rather than purely instance-specific circuits.

These findings point to circuit reuse as a promising direction for reducing discovery overhead. Rather than rediscovering each circuit from scratch, future work could cache representative circuits and reuse them as priors or initialization for subsequent edits, while retaining instance-specific refinement when needed. Such reuse could amortize circuit discovery without abandoning the fine-grained structure that motivates circuit-based editing.


\section{Related Work}
\subsection{Knowledge Editing for LLMs}
Knowledge Editing~\cite{wang2024knowledge,zhang2024comprehensive,wang2024easyedit} aims to update factual associations in Large Language Models (LLMs) without the computational cost of retraining or the degradation associated with catastrophic forgetting. Methodologies in this field have evolved from manipulating static parameter storage to aligning dynamic pathways.
Foundational approaches treat Transformer Feed-Forward Networks (FFNs) as key-value associative memories. 
For instance, ROME~\cite{meng2022locating} employs causal tracing to localize factual storage to specific mid-layer MLPs and formulates a rank-one update to insert a new fact tuple.
While ROME is effective for single edits, it struggles with scalability. MEMIT~\cite{mengmass} addresses this by distributing updates across a range of critical layers. By solving a least-squares problem with residual spreading, MEMIT allows for the simultaneous injection of thousands of memories, significantly outperforming ROME in batch-editing scenarios.
In sequential "lifelong" editing scenarios, maintaining the stability of previously learned information is paramount. WISE~\cite{wang2024wise} critiques the destructive nature of continuous parameter overwriting and introduces a side memory architecture coupled with a routing mechanism.
Conversely, AlphaEdit~\cite{fangalphaedit} enforces stability through geometric constraints rather than architectural changes. It projects parameter perturbations onto the Null Space of the preserved knowledge matrix to ensure that the model's performance on pre-existing data remains invariant.
Another critical limitation of storage-focused editors is the "Reasoning Gap", where models recall a specific fact but fail to apply it in multi-step inference. IFMET~\cite{zhanglocate} identifies a functional stratification: shallow layers store explicit facts, while deep layers process implicit subjects necessary for multi-hop reasoning. It proposes a two-stage strategy that edits both shallow and deep layers to ensure the updated knowledge propagates through the entire reasoning chain.
On the other hand, CaKE~\cite{yao2025cake} moves beyond the storage metaphor entirely. It posits that knowledge is utilized via dynamic reasoning circuits. CaKE generates complex, circuit-aware training data designed to activate these specific pathways, stimulating the model to functionally integrate new information into its logical processing.

\subsection{Mechanistic Interpretability}
Mechanistic interpretability~\cite{rai2024practical,bereska2024mechanistic,sharkey2025open} aims to reverse-engineer model behaviors into human-understandable algorithms implemented by specific subgraphs or "circuits".
Pioneering work manually identified circuits for tasks like Indirect Object Identification~\cite{wanginterpretability} and arithmetic~\cite{hanna2023does}. To scale this analysis, automated methods were developed: ACDC~\cite{conmy2023towards} employs activation patching to prune irrelevant edges but is computationally expensive. Edge Attribution Patching (EAP)~\cite{nanda2023attribution} approximates these effects via gradients for efficiency. Most recently, EAP with Integrated Gradients (EAP-IG)~\cite{hanna2024have} was proposed to resolve the gradient saturation issue in EAP, offering a superior balance of speed and faithfulness. Our work leverages EAP-IG to identify dynamic reasoning circuits for knowledge editing, moving beyond the static localization assumptions of prior editing methods like ROME~\cite{zhanglocate} or MEMIT~\cite{mengmass}.

\section{Conclusion}
We introduce MCircKE, a mechanistically grounded framework for knowledge editing that addresses the reasoning gap between single-hop factual recall and multi-hop inference.
Our analysis shows that multi-hop reasoning relies on sparse, distributed circuitry beyond direct factual recall, with distinct contributions to intermediate information propagation and final prediction.
To address this, MCircKE dynamically maps these transient reasoning pathways and performs surgical interventions within the identified circuits. Extensive evaluations demonstrate the effectiveness of MCircKE in multi-hop factual recall for knowledge editing in LLMs, highlighting circuit structure as a promising basis for precise and generalizable knowledge editing.

\section*{Limitations}
\paragraph{Sequential Editing and Interference}
Our current evaluation focuses on single edits. It remains an open question whether the proposed framework has sufficient capacity to support thousands of sequential edits without saturation or cross-talk (interference between different edited facts sharing the same routing heads).


\paragraph{Circuit Discovery Cost and Trade-off}
Calculating EAP-IG requires computing gradients for every edge, which, while efficient compared to activation patching, still incurs a computational overhead ($O(m)$ backward passes) that may be prohibitive for real-time editing of extremely large models.
Besides, EAP-IG serves as a gradient-based approximation to exact activation patching. The fidelity of this approximation is governed by the number of integration steps $m$. While reducing $m$ improves computational efficiency, it may hinder convergence to the true attribution scores, resulting in the identification of incomplete circuits that do not faithfully represent the model's reasoning topology.

\paragraph{Threshold Sensitivity}
The top-$K$ pruning strategy imposes a fixed sparsity level across all edits. This heuristic may be suboptimal because complex facts may require larger circuits while simple facts may require smaller ones, and these requirements can also vary across model scales, necessitating future work on adaptive thresholding.

\bibliography{custom}

\appendix

\section{Notation}\label{app:notation}

\resizebox{\columnwidth}{!}{%
\begin{tabular}{ll}
\toprule
Symbol & Meaning \\
\midrule
$\mathcal{G}=(\mathcal{V},\mathcal{E})$ & Full model computational graph \\
$\mathcal{C}(K) \subseteq \mathcal{G}$ & Causal circuit: top-$K$ EAP-IG edges (Eq.~\ref{eq:circuit-def}) \\
$\phi(e)$ & EAP-IG attribution score of edge $e$ (Eq.~\ref{eq:approx}) \\
$I_{\mathrm{clean}}$ & Clean prompt eliciting target behavior \\
$I_{\mathrm{corrupt}}$ & Structurally matched contrastive prompt \\
$m$ & IG Riemann-sum steps (set to $5$) \\
$K$ & Top-$K$ circuit-size pruning threshold \\
$r$, $\alpha$ & LoRA rank, scaling factor \\
M-Acc & Multi-hop Accuracy \\
S-Acc & Single-hop Accuracy \\
$\Delta W$ & Parameter update applied by an editing method \\
\bottomrule
\end{tabular}
}

\section{Prerequisite}
\subsection{LLM as a Computational Graph}\label{app:eapig}
We conceptualize the LLM as a Directed Acyclic Graph (DAG), denoted as $\mathcal{G} = (\mathcal{V}, \mathcal{E})$.
In this representation, the nodes $\mathcal{V}$ are the specific computational sub-modules of the network, and the edges $\mathcal{E}$ represent the flow of information between them via the residual stream.
\paragraph{Nodes ($\mathcal{V}$)} 
To enable fine-grained editing, we decompose the standard Transformer blocks into the atomic linear projections of the models:
\begin{equation}
\resizebox{\linewidth}{!}{%
  \ensuremath{
    \mathcal{V} = \{ W_Q^{(l,h)}, W_K^{(l,h)}, W_V^{(l,h)}, W_O^{(l,h)}, W_{MLP}^{(l)} \mid l \in [1, L], h \in [1, H] \}
  }%
}
\end{equation}
where $L$ is the number of layers and $H$ is the number of heads per layer. $W_Q$, $W_K$, $W_V$ represent the query, key, and value projections for the $h$-th head in layer $l$, $W_O$ is the output projection, and $W_{MLP}$ represents the two-layer feed-forward network.

\paragraph{Edges ($\mathcal{E}$)}
An edge $e_{u \to v}$ exists if the output of module $u$ contributes to the input of module $v$ via the residual stream.

\subsection{Edge Attribution Patching with Integrated Gradients (EAP-IG)}\label{app:eapig2}
To identify the specific subgraph responsible for a model's behavior, we require a method to attribute the model's output to specific internal edges. Standard methods like Activation Patching~\cite{meng2022locating} are faithful but computationally expensive, while vanilla gradient attribution suffers from saturation. To balance fidelity and efficiency, we employ Edge Attribution Patching with Integrated Gradients (EAP-IG)~\cite{hanna2024have}.

Specifically, vanilla EAP~\cite{nanda2023attribution} approximates the importance of an edge $e$ by computing the product of the edge's activation $x_e$ and the gradient of the loss with respect to that activation $\frac{\partial \mathcal{L}}{\partial x_e}$:
\begin{equation}
    \phi_{EAP} = x_e\cdot \frac{\partial \mathcal{L}}{\partial x_e}
\end{equation}
While requiring only one backward pass, this method fails in deep non-linear networks due to the saturation effect. When a neuron is fully activated, the local gradient is near zero, even if that neuron is critically important for the output. This often yields "broken" circuits that miss key intermediate nodes.

To overcome saturation, EAP-IG accumulates gradients along a linear path between a corrupted baseline input ($I_{corrupt}$) and the clean target input ($I_{clean}$).
Let $x_e$ denote the activation along edge $e$. A path is defined as $x_e(\gamma) = x_e^{corrupt} + \gamma(x_e^{clean} - x_e^{corrupt})$ for $\gamma \in [0,1]$.
The attribution score $\phi(e)$ for an edge $e$ is calculated as:
\begin{equation}
    \phi(e) = (x_e^{clean} - x_e^{corrupt}) \times \int_{\gamma=0}^{1} \frac{\partial \mathcal{L}(x(\gamma))}{\partial x_e(\gamma)} d\gamma
\end{equation}
In practice, this integral is approximated using a Riemann sum with $m$ steps:
\begin{equation}\label{eq:approx2}
    \phi(e) \approx (x_e^{clean} - x_e^{corrupt}) \times \frac{1}{m} \sum_{k=1}^{m} \frac{\partial \mathcal{L}(x(\frac{k}{m}))}{\partial x_e}
\end{equation}
This formalism provides a robust, continuous estimate of causal importance, enabling us to map the full reasoning circuitry even through saturated MLP layers.


\section{Additional Details on Functional Role Characterization}
\label{app:mech_func}
Here we provide additional details for the analysis in Section~\ref{ana3}.
\subsection{Causal Characterization of Component Roles}
For each two-hop reasoning instance $s \rightarrow b \rightarrow o$, we construct a clean input ($I_{\mathrm{clean}}$) and a structurally matched corrupted input ($I_{\mathrm{corrupt}}$). The corrupted prompt preserves the template and relation structure of the clean prompt while replacing the subject entity, thereby changing both the intermediate bridge entity and the final answer. 
We evaluate two complementary targets. The \underline{bridge} target uses the first-hop query and contrasts the clean bridge entity $b$ with its corrupted counterpart $b'$. The \underline{final} target uses the complete multi-hop query and contrasts the clean final answer $o$ with the corrupted answer $o'$. The bridge entity is therefore an intermediate variable in the reasoning chain rather than an entity explicitly provided by the final multi-hop question.
The clean and corrupted sequences are constrained to have the same token length, which allows component activations to be restored position-wise between the two forward passes.

We evaluate each component with respect to two stages of the reasoning chain: recovering the intermediate bridge entity and predicting the final answer. For each target $t\in\{\mathrm{bridge},\mathrm{final}\}$, we measure model behavior using the logit difference between the clean and corrupted target tokens at the final prediction position: \begin{equation} m_t(x) = \operatorname{logit}_{x}[-1,A_t] - \operatorname{logit}_{x}[-1,B_t], \label{eq:role-logit-diff} \end{equation} where $A_t$ and $B_t$ denote the clean and corrupted target tokens, respectively. Specifically,
\begin{equation}
\resizebox{0.98\columnwidth}{!}{$
(A_{\mathrm{bridge}},B_{\mathrm{bridge}}) = (b,b'),
\quad
(A_{\mathrm{final}},B_{\mathrm{final}}) = (o,o')
$}
\label{eq:role-targets}
\end{equation}
Thus, $m_{\mathrm{bridge}}$ measures the model's preference for the correct intermediate entity $b$ over its corrupted counterpart $b'$, while $m_{\mathrm{final}}$ measures the corresponding preference for the correct final answer $o$ over $o'$.
For each target, the total effect of the corruption is defined as \begin{equation} \Delta_t = m_t(I_{\mathrm{clean}}^{t}) - m_t(I_{\mathrm{corrupt}}^{t}). \label{eq:role-total-effect} \end{equation}
This quantity provides the normalization baseline for the component-level intervention described below. For multi-token bridge entities, we use the first token of the entity; $b'$ is the bridge entity induced by the corrupted prompt rather than a randomly selected token.

\paragraph{Component-level causal intervention.}
We next measure how much of the clean--corrupt behavioral difference can be recovered by restoring a single model component. For an attention head $(\ell,h)$, let
$$
z_{\ell,h}(x)\in\mathbb{R}^{T\times d_{\mathrm{model}}}
$$
denote its output after the head-specific output projection and before addition to the residual stream.
During the corrupted forward pass, we replace the output of a single component $c$ with its activation from the corresponding clean run while leaving all other activations unchanged:
\begin{equation}
\widetilde{M}_{c}^{\,t}
=
M\left(
I_{\mathrm{corrupt}}^{t}
\;\middle|\;
\operatorname{do}
\left[
z_c
\leftarrow
z_c(I_{\mathrm{clean}}^{t})
\right]
\right).
\label{eq:component-restoration}
\end{equation}

For attention heads, the complete $T \times d_{\mathrm{model}}$ output is restored across all token positions rather than only at the final position. We apply the same intervention principle to MLP components.

\paragraph{Normalized indirect effect.}
For each component $c$ and target $t$, we define its normalized indirect effect as the fraction of the total clean--corrupt difference recovered by restoring that component alone:

\begin{equation}
\mathrm{IE}_{t}(c)
=
\frac{
m_t(\widetilde{M}_{c}^{\,t})
-
m_t(I_{\mathrm{corrupt}}^{t})
}{
m_t(I_{\mathrm{clean}}^{t})
-
m_t(I_{\mathrm{corrupt}}^{t})
}.
\label{eq:normalized-ie}
\end{equation}

Accordingly, $\mathrm{IE}_{t}(c)=1$ indicates that restoring component $c$ recovers the full clean--corrupt effect for target $t$, whereas $\mathrm{IE}_{t}(c)=0$ indicates no recovery. Negative values indicate that restoring the component moves the prediction further away from the clean behavior.

For every component, we compute two effects:
\begin{equation}
\resizebox{0.95\columnwidth}{!}{$
\mathrm{IE}_{\mathrm{bridge}}(c)
=
\frac{
m_{\mathrm{bridge}}
\left(
\widetilde{M}_{c}^{\,\mathrm{bridge}}
\right)
-
m_{\mathrm{bridge}}
\left(
I_{\mathrm{corrupt}}^{\mathrm{bridge}}
\right)
}{
m_{\mathrm{bridge}}
\left(
I_{\mathrm{clean}}^{\mathrm{bridge}}
\right)
-
m_{\mathrm{bridge}}
\left(
I_{\mathrm{corrupt}}^{\mathrm{bridge}}
\right)
},
$}
\label{eq:ie-bridge}
\end{equation}
and

\begin{equation}
\resizebox{0.95\columnwidth}{!}{$
\mathrm{IE}_{\mathrm{final}}(c)
=
\frac{
m_{\mathrm{final}}
\left(
\widetilde{M}_{c}^{\,\mathrm{final}}
\right)
-
m_{\mathrm{final}}
\left(
I_{\mathrm{corrupt}}^{\mathrm{final}}
\right)
}{
m_{\mathrm{final}}
\left(
I_{\mathrm{clean}}^{\mathrm{final}}
\right)
-
m_{\mathrm{final}}
\left(
I_{\mathrm{corrupt}}^{\mathrm{final}}
\right)
}.
$}
\label{eq:ie-final}
\end{equation}
We average these effects over $N=100$ cases:
\begin{equation}
\overline{\mathrm{IE}}_{t}(c)
=
\frac{1}{N}
\sum_{n=1}^{N}
\mathrm{IE}_{t}^{(n)}(c),
\label{eq:mean-ie}
\end{equation}
We evaluate all 720 attention heads and 36 MLP modules in GPT-2 Large for both targets.

\paragraph{Functional-role assignment.}
Figure~\ref{fig:panel}(d) plots
$\overline{\mathrm{IE}}_{\mathrm{bridge}}$
against
$\overline{\mathrm{IE}}_{\mathrm{final}}$
for components identified by EAP-IG as circuit members.
Because relative comparisons between two near-zero effects are not meaningful, we assign a functional role only when
\begin{equation}
\max
\left(
\overline{\mathrm{IE}}_{\mathrm{bridge}}(c),
\overline{\mathrm{IE}}_{\mathrm{final}}(c)
\right)
\geq 0.01.
\label{eq:ie-threshold}
\end{equation}

Components passing this threshold are categorized as

\begin{equation}
\resizebox{0.95\columnwidth}{!}{$
c \in
\begin{cases}
\text{bridge-dominant},
&
\overline{\mathrm{IE}}_{\mathrm{bridge}}(c)
>
2\,\overline{\mathrm{IE}}_{\mathrm{final}}(c),
\\[3pt]
\text{final-dominant},
&
\overline{\mathrm{IE}}_{\mathrm{final}}(c)
>
2\,\overline{\mathrm{IE}}_{\mathrm{bridge}}(c),
\\[3pt]
\text{both},
&
\text{otherwise}.
\end{cases}
$}
\label{eq:role-assignment}
\end{equation}

Components below the $1\%$ threshold are shown separately as having negligible measured effects. We use inequalities rather than an effect ratio so that the criterion remains well defined when one of the two effects is negative.

The resulting distribution shows that the two effects are not mutually exclusive: many components contribute to both targets, while distinct bridge-dominant and final-dominant groups remain identifiable. We therefore interpret these labels as {causal functional biases}, rather than as a strict decomposition of the network into disjoint functional modules.

\subsection{Layer Distribution of Attention-Head Roles}
We further examine whether these causal biases exhibit systematic structure across network depth. This analysis considers the 54 attention heads whose maximum effect satisfies the $0.01$ threshold, comprising 14 bridge-dominant, 31 shared, and 9 final-dominant heads.

For each role $q$, we estimate the distribution of its layer indices using a Gaussian kernel density estimator:
\begin{equation}\resizebox{0.95\columnwidth}{!}{$
\hat{f}_{q}(x)
=
\frac{1}{n_q h \sqrt{2\pi}}
\sum_{i \in q}
\exp
\left(
-\frac{(x-L_i)^2}{2h^2}
\right),
\quad
h = 2.2,
$}
\label{eq:role-kde}
\end{equation}
where $L_i$ is the layer index of head $i$ and $n_q$ is the number of heads assigned to role $q$. Each density curve in Figure~\ref{fig:headrole} is normalized to its own maximum; the curves therefore represent the shape and location of each depth distribution.

To quantify the depth shift between bridge-dominant and final-dominant heads, we use the difference in median layer as the test statistic:

\begin{equation}
T_{\mathrm{obs}}
=
\operatorname{median}
\left(
L_{\mathrm{final}}
\right)
-
\operatorname{median}
\left(
L_{\mathrm{bridge}}
\right).
\label{eq:median-depth}
\end{equation}
Under the null hypothesis that the two role labels are exchangeable with respect to network depth, we pool their layer indices and randomly reassign them while preserving group sizes. The two-sided permutation $p$-value is computed as

\begin{equation}
p
=
\frac{
\#\left\{
\left|T_{\mathrm{perm}}\right|
\geq
\left|T_{\mathrm{obs}}\right|
\right\}
}{
N_{\mathrm{perm}}
}.
\label{eq:permutation-test}
\end{equation}
Bridge-dominant heads have a median layer of 23, whereas final-dominant heads have a median layer of 31, yielding an eight-layer shift with $p=0.016$. We interpret this as a systematic depth shift rather than a clean separation between the two roles.


Together, the intervention and depth analyses indicate that intermediate bridge propagation and final-answer prediction are causally distinguishable but distributed computations. Their substantial overlap argues against a rigid layer-wise decomposition, while the systematic depth shift provides complementary evidence that multi-hop reasoning involves information-propagation mechanisms beyond those required for direct factual recall.




\section{Additional Experimental Results}
\label{app:exp_more}
Additional results are presented in Table~\ref{tab:qwen}, Table~\ref{tab:xl3k} and Table~\ref{tab:large}.

\begin{table}[h]
\centering
\begin{tabular}{l c c}
\toprule
Method & M-Acc & S-Acc \\
\midrule
LoRA    & 31.23 & 35.87 \\
ROME     & 39.03 & 85.77 \\
CaKE  & 35.40 & 84.53 \\
\rowcolor{lightgray}
\textbf{MCircKE} & \textbf{46.20} & \textbf{87.93} \\
\bottomrule
\end{tabular}
\caption{Results on Qwen2.5-7b / MQuAKE-CF-3K-v2.}\label{tab:qwen}
\end{table}

\begin{table}[h]
\centering
\begin{tabular}{l c c}
\toprule
Method & M-Acc & S-Acc \\
\midrule
LoRA    & 17.47 & 20.25 \\
ROME     & 16.10 & 60.80 \\
CaKE  & 47.73 & 74.90 \\
\rowcolor{lightgray}
\textbf{MCircKE} & \textbf{55.50} & \textbf{78.80} \\
\bottomrule
\end{tabular}
\caption{Results on GPT-2 XL / MQuAKE-CF-3K.}\label{tab:xl3k}
\end{table}

\begin{table}[h]
\centering
\begin{tabular}{l c c}
\toprule
 & MQuAKE-CF-3K & MQuAKE-T \\
\midrule
LoRA    & 10.39 &  19.39\\
ROME     & 17.53 & 5.19 \\
CaKE  & 34.97 & 40.29 \\
\rowcolor{lightgray}
\textbf{MCircKE} & \textbf{47.02} & \textbf{98.94} \\
\bottomrule
\end{tabular}
\caption{Results on GPT-2 Large (M-Acc).}\label{tab:large}
\end{table}

\section{Auxiliary}
\label{sec:appendix}
\subsection{Licenses}
This study utilizes several open-source artifacts, all of which are distributed under permissive licenses compatible with academic research.
\begin{itemize}
    \item MQuAKE-3K Dataset: Distributed under the MIT License.
    \item GPT-2 (Large/XL): Released by OpenAI under the Modified MIT License.
    \item GPT-J (6B): Released by EleutherAI under the Apache-2.0 License.
    \item CaKE (Baseline): The official implementation and data are distributed under the MIT License.
    \item ROME/MEMIT/AlphaEdit (Baselines): The official codebases for these methods are distributed under the MIT License.
\end{itemize}

\subsection{Intended Use Consistency}
All artifacts employed in this study were used in a manner consistent with their intended purposes as defined by their creators.
\begin{itemize}
    \item MQuAKE-3K: This dataset was specifically designed to evaluate the multi-hop reasoning capabilities of edited language models. Our usage strictly adheres to this evaluation protocol.
    \item Language Models (GPT-2, GPT-J): These models are intended for research into the properties of large language models, including interpretability and fine-tuning. Our work, which investigates the internal mechanisms of these models via EAP-IG and applies circuit-based editing, aligns with this research scope.
    \item Synthetic Data: The synthetic prompts used for the MCircKE were generated using ChatGPT (OpenAI). This usage falls within the acceptable use policy for research purposes, specifically for generating non-sensitive, factual reasoning chains.
\end{itemize}

\subsection{Documentation of Artifacts}
\textbf{Language \& Domain}: All models and datasets used in this study process English language text. The domain of the MQuAKE-3K dataset is factual knowledge graphs (derived from Wikidata), covering entities such as people, organizations, locations, and creative works.

\noindent
\textbf{Dataset Characteristics}: The MQuAKE-3K dataset consists of 3,000 counterfactual editing instances. Each instance includes a base edited fact and a set of multi-hop questions (2-hop, 3-hop, 4-hop) that test the model's ability to propagate this edit.

\noindent
\textbf{Model Specifications}:
GPT-2 Large - 774M parameters, 36 layers, 1280 hidden dimension; GPT-2 XL - 1.5B parameters, 48 layers, 1600 hidden dimension; GPT-J - 6B parameters, 28 layers, 4096 hidden dimension

\noindent
\textbf{Demographics}: The datasets are derived from Wikipedia and Wikidata, which may contain inherent biases reflecting the demographics of English-speaking internet users. No human subjects were involved in this specific study.


\end{document}